# Highlights

**Supporting Automated Fact-checking across Topics:**
**Similarity-driven Gradual Topic Learning for Claim Detection**

Amani S. Abumansour, Arkaitz Zubiaga

- We investigate check-worthy claim detection to support automated fact-checking across topics.
- We propose Gradual Topic Learning (GTL), a gradual learning strategy to adapt a claim detection model to new topics.
- We further propose Similarity-driven Gradual Topic Learning (SGTL), which incorporates a similarity-based strategy on GTL for improved gradual learning.
- Our experiments demonstrate the effectiveness of GTL and SGTL, and we highlight future avenues to tackle open challenges.

# Supporting Automated Fact-checking across Topics: Similarity-driven Gradual Topic Learning for Claim Detection

Amani S. Abumansour[a,b,<], Arkaitz Zubiaga[a]

[a]*Queen Mary University of London, United Kingdom*
[b]*Taif University, Saudi Arabia*

**Abstract**

Selecting check-worthy claims for fact-checking is considered a crucial part of expediting the fact-checking process by filtering out and ranking the check-worthy claims for being validated among the impressive amount of claims could be found online. The check-worthy claim detection task, however, becomes more challenging when the model needs to deal with new topics that differ from those seen earlier. In this study, we propose a domain-adaptation framework for check-worthy claims detection across topics for the Arabic language to adopt a new topic, mimicking a real-life scenario of the daily emergence of events worldwide. We propose the Gradual Topic Learning (GTL) model, which builds an ability to learning gradually and emphasizes the check-worthy claims for the target topic during several stages of the learning process. In addition, we introduce the Similarity-driven Gradual Topic Learning (SGTL) model that synthesizes gradual learning with a similarity-based strategy for the target topic. Our experiments demonstrate the effectiveness of our proposed model, showing an overall tendency for improving performance over the state-of-the-art baseline across 11 out of the 14 topics under study.

## 1. Introduction

Online platforms such as social media have become common sources of information and news for an increasing number of people worldwide (Zubiaga, 2019), which are also however known to be rife with misinformation (Fernandez and Alani, 2018). This has increased the need to develop automated methods to support fact-checking, or at least parts of the fact-checking pipeline to alleviate the otherwise burdensome human endeavour (Zeng et al., 2021). The first component in an automated fact-checking pipeline is typically the claim detection component (Elsayed et al., 2019; Panchendrarajan and Zubiaga, 2024), which consists in determining which sentences, out of a large collection, constitute claims that are worthy of fact-checking and hence should be considered in subsequent steps of the fact-checking pipeline to be checked against pieces of evidence. Despite the importance of the claim detection component in the fact-checking pipeline, it has received considerably less attention than the latter claim verification component.

Research in claim detection has recently increased in popularity (Panchendrarajan and Zubiaga, 2024), arguably motivated by a series of related shared tasks under the umbrella of the CheckThat! lab (Hasanain et al., 2024), which has consistently had a subtask dealing with check-worthy claim detection across the years (Elsayed et al., 2019), as well as due to the need to develop complete fact-checking pipelines. Despite this increasing interest in the task, research has overlooked the realistic scenario where a claim detection model trained on data pertaining to a set of topics needs to deal with the detection of claims in new, unseen topics. Since new emerging topics have a high potential to occur in a real-world scenario, it is essential to consider the difficulty of generalising the model's ability to detect claims beyond the data that resembles what has been seen during training by adapting it to tackle new, unseen topics.

In an earlier study, Abumansour and Zubiaga (2023) presented what was, to the best of our knowledge, the first study on cross-topic claim detection. In this study, they introduced the AraCWA model which achieved improved performance across unseen topics by leveraging data augmentation strategies. However,

[<]Corresponding author
`a.s.a.abumansour@qmul.ac.uk` (A.S. Abumansour)
ORCID(s): 0009-0003-5868-4898 (A.S. Abumansour); 0000-0003-4583-3623 (A. Zubiaga)

this study also highlighted the challenging nature of the cross-topic claim detection task, with room for improvement across a range of the topics. As the best-performing model for the task to date, here we use AraCWA as our main baseline model.

In the present study, we are interested in exploring the potential of transfer learning strategies to support the cross-topic claim detection task. To advance research in this direction, we embrace a recent line of research based on gradual learning (Soviany et al., 2022), i.e. where a model is shown training examples of increasing difficulty so that it can gradually adapt to the challenges of the new topic not previously known to the model. With this work, we are the first to adapt gradual learning for claim detection, offering a novel perspective of the learning approach in machine learning that mirrors the human learning process in assimilating new information, in turn supporting the fact-checking process on new topics.

To achieve this objective, we propose the GTL (Gradual Topic Learning) model and the SGTL (Similarity-driven Gradual Topic Learning) model, two variants of a novel claim detection model that incorporates a gradual learning strategy so the model can adapt to new topics. While GTL provides the framework for gradually learning the properties of claims for topics that differ from those seen during training, SGTL further enhances the framework with a similarity-based strategy to more carefully select the samples to be gradually incorporated into the training.

We test GTL and SGTL on a dataset involving 14 different topics pertaining to the CheckThat! labs, studying the overall performance of the model as well as a more fine-grained study on the generalisability of the model across topics. Our experiments show that the SGTL model we developed exhibits superior performance and exceptional capability in detecting check-worthy claims, attaining a 9% average improvement across the 14 topics under study compared to the baseline. In addition to outperforming the baseline on average, this performance is reflected individually in 11 of the 14 topics, with comparable performance in 2 of the remainder topics, and decreased performance in one topic. While we study the effectiveness of GTL and SGTL in the Arabic language and leveraging AraBERT as the underlying Large Language Model (LLM), the model is extensible to other languages and to other models.

The remainder of this paper is organized as follows. Section 2 reviews the related work and presents the background. Section 3 describes the problem formulation for adapting a new topic to the claim detection model. Section 4 discusses the datasets used for our experiments. Section 5 introduces the methodology of our research, including our proposed GTL and SGTL models as well as the experimental setup. Section 6 presents and analyzes the results of our experiments, delving into the different topics under study and performing a qualitative analysis comparing with the baseline model, including a discussion of the implications of our results. Finally, Section 7 concludes the paper, followed by the limitations of this study and potential directions for future research.

## 2. Related Work

An automated fact-checking system is a critical tool to assist in the challenge of verifying the numerous claims manifesting in social media, which are likely to contain a high volume of misinformation that must be debunked (Babakar and Moy, 2016). These efforts are currently spearheaded by globally renowned organizations such as Full Fact, Chequeado, and Neutral, along with academic NLP-based efforts to support scaling up and supporting with partial automation of the process to support fact-checkers (Nakov et al., 2021a).

In what follows, we discuss two lines of work related to our research, including check-worthy claim detection and gradual learning.

### 2.1. Checkworthy claim detection

Within the task of automated fact-checking, claim check-worthiness detection is a key component that prioritises important claims requiring fact-checking, where a check-worthy claim is defined as a sentence containing a verifiable factual statement (Hassan et al., 2017a). The task has seen significant progression over the last decade, with early approaches using generative and discriminative classification models, to more recent studies leveraging Transformers and other kinds of LLMs.

***Early approaches to claim detection.*** As one of the early efforts in this direction, Hassan et al. (2017b) defined a "claim spotter" component as a part of the first automated fact-checking system named

"ClaimBuster," which seeks to identify the check-worthy claims from US presidential election debates. A slightly different approach used discourse and context features to enhance the detection of check-worthy claims from political debates (Atanasova et al., 2019). In another research endeavor, Konstantinovskiy et al. (2021) created the CNC model, which uses named entities, part-of-speech tags, and InferSent embeddings to identify claims.

***Claim detection in the Arabic language.*** In the context of the Arabic language, one of the main challenges has been the lack of suitable datasets annotated in Arabic. To overcome this limitation, Arabic translations of original English sentences from political debates were used by ClaimRank to train the neural network model (Jaradat et al., 2018). At CheckThat! 2018, models such as Gradient boosting and k-nearest neighbors were examined to estimate the check-worthy claims from five debates of the US Presidential in 2016 and Donald Trump's acceptance speech for the Arabic check-worthiness challenge (Atanasova et al., 2018). However, prior studies were limited to political discussions and did not include for example social media posts which have become widely studied in recent years. Moreover, traditional machine-learning models were examined at that time. The dynamics started to shift and appeared in the most recent series of CheckThat! shared tasks when datasets were released, including tweets covering topics other than politics that coincided with the COVID-19 pandemic and several other events that occurred during that period. These also included tweets in other languages, including in Arabic. Consequently, claim detection models for the Arabic language used Transformers like AraBERT and BERT-Base-Arabic, as well as natively multilingual models like mBERT.

***Recent trends in claim deetection.*** The advent of the CheckThat! series of tasks in fact-checking, and particularly in claim detection, showed a shift in the types of models used. Notably, an increasing number of methods started to leverage transformer models (Hasanain et al., 2020; Shaar et al., 2021). For instance, in the 2020 edition of Checkthat!, Williams et al. (2020) obtained the highest MAP score, reaching about 30% over baselines by fine-tuning the pre-trained language model, namely AraBERT. In the following edition, Abumansour et al. (2021) achieved the joint fourth rank by conducting extra preprocessing techniques for the training data and incorporating additional data to fine-tune the AraBERT model. Later, in the edition of CheckThank! 2022, the top rank was for a model that utilized data with multiple languages trained for the multilingual transformer, such as mT5 (Du et al., 2022; Nakov et al., 2022).

More recently, research in claim detection witnessed a growth towards the development of Multimodal and Multi-Genre models that consider detecting check-worthy claims from tweets, including text that appears with its corresponding image or independent text taken from several sources, including tweets, political debates, and political speeches As observed in the shared tasks given by CheckThat! Labs for 2023 and 2024 (Alam et al., 2023; Hasanain et al., 2024). Aziz et al. (2023) developed a multimodal method that stands as the top performer, which was attained by uniting the visual and textual representations using ConvNext for extracting image features and AraBert to represent the Arabic tweet texts with a BiLSTM on top to address the long-term contextual features. For Multigenre models, Sadouk et al. (2023) fine-tuned MARBERT after reducing the size of the majority class to tackle the unbalanced training set. In a similar approach, Aarnes et al. (2024) leveraged multilingual datasets by evaluating a variety of Large Language Models (LLM), in particular GPT-4, Mistral 7b, XLMR, the fine-tuned XLMR, GPT-3.5, fine-tuned GPT-3.5 which the latter achieved the best performance, aiming to improve check-worthiness detection's generalizability.

***Gap in claim detection across topics.*** A more comprehensive discussion of methods to claim detection has been discussed by Zeng et al. (2021); Panchendrarajan and Zubiaga (2024). However, there is still a dearth of research in cross-topic claim detection, which we study in our work by proposing a novel method leveraging gradual learning to enable the model to gradually adapt to new, unseen topics. The only such effort, to the best of our knowledge, has been by the introduction of the AraCWA model (Abumansour and Zubiaga, 2023), a model that builds on data augmentation to further the generalisability of the model across topics. Further, in the present work we propose a strategy that leverages a gradual learning strategy to improve the ability to generalise across topics, including over AraCWA which we use as a baseline model.

## 2.2. Domain Adaptation learning techniques

In the realm of machine learning specifically, transfer learning offers a range of methods, including domain adaptation. This method intends to mitigate the effects of domain shift and strive to achieve optimal performance when the model is trained on the source domain and tested on the target domain, where both domains are distinct but share some commonalities (Singhal et al., 2023).

Domain adaptation is widely applied to various fields, such as computer vision, speech recognition, machine translation, and natural language processing. One learning technique developed for domain adaptation is "Gradual learning," which is recommended for cases when training data could be incrementally available over time (Hou et al., 2019), which is similar to our task when there are unprecedented topics emerge. (Xu et al., 2021) experimented with gradual learning by fine-tuning the model for multiple stages for two NLP tasks, dialogue state tracking and event extraction tasks, where the performance of multiple stages of gradual fine-tuning conquered standard single-stage fine-tuning in both tasks. A related direction that mimics the human education system and can be leveraged for domain adaptation is curriculum learning. It was first introduced in machine learning by Bengio et al. (2009), where the key concept is based on structured sequencing of information by starting with easy examples and gradually increasing the difficulty while progressing in the model's training. This yields smooth convergence between examples and leads to improved model generalization. Similarly, Zhang et al. (2019) applied a curriculum learning for domain adaptation to a Neural Machine Translation model. Despite their broad use across different tasks and applications, these learning techniques have not been experimented with for check-worthy claim detection models.

## 3. Problem Formulation

We next describe the main concepts relevant to our problem, including a formulation of the claim detection task, as well as key elements in a gradual learning strategy.

### 3.1. Claim Detection

The check-worthy claim detection task consists in determining which sentences constitutes claims that are worthy of being fact-checked. As such, the task takes a collection of sentences S = {$s_1$, $s_2$,..., $s_n$} as input and a model needs to determine which of the sentences are worthy of, and therefore should be prioritized for, fact-checking.

While the task is sometimes formalized as a classification problem where the model makes a binary decision on each sentence determining whether it is check-worthy or not, here we adopt the widely used formulation as a ranking task. Where the check-worthy claim detection task is formulated as a ranking problem, the task consists in ranking the input sentences based on their likelihood to be claims that are check-worthy. The output of the ranking model will then be an ordered list of sentences, where a model that puts more check-worthy claims in the top positions will be deemed as the best performing.

### 3.2. Gradual Learning

The concept of gradual learning relies on employing scheduled multiple stages by applying several criteria (Soviany et al., 2022). We next describe the different elements needed in the development of a gradual learning pipeline in the context of check-worthy claim detection.

#### 3.2.1. Source and Target Domains

In our experiments, we separate the topics within the dataset such that one of the topics is left as the target domain each time, whereas the remainder of the topics are considered to be the source domain.

***Source Domain*** The source domain consists of all but one of the topics in the dataset, which is used for training the model:

$$\mathcal{D}_S = \{(x_i^S, y_i^S)\}_{i=1}^{n_S}$$

where $x_i^S$ represents a sentence of check-worthy claims in the source topics. $y_i^S$ represents the corresponding labels e.g. {CW, NCW}, where CW = check-worthy and NCW = not check-worthy. $n_S$ is the number of examples in the source topics.

***Target Domain*** The target domain consists of a single topic, which has been left out of the source domain and therefore from the training process, and is used for testing:

$$\mathcal{D}_T = \{(x_i^T, y_i^T)\}_{i=1}^{n_T}$$

where $x_i^T$ represents the check-worthy claims sentence that should be fact-checked points in the target top. $y_i^T$ represents the corresponding labels either {CW, NCW}, where CW = check-worthy and NCW = not check-worthy. $n_T$ is the number of examples in the target topic.

***Feature space and label space.*** The feature space $\mathcal{X}$, and label space $\mathcal{Y}$ are the same for both domains:

$$\mathcal{X}_S = \mathcal{X}_T = \mathcal{X}, \qquad \mathcal{Y}_S = \mathcal{Y}_T = \mathcal{Y}$$

### 3.2.2. Few-shot Learning

We employ the gradual learning by leveraging the AraCW model for detecting check-worthy claims, which mainly depends on a few-shot learning settingAbumansour et al. (2021). The few-shot learning is an approach in machine learning that implicates learning knowledge from different distributions of tasks to adapt to new tasks. The purpose is to empower the model to generalize these examples and gain the capability to render precise predictions on new or unobserved data previously. Few-shot learning is advantageous when data is limited or requires time and labor to annotate extra data (Brown, 2020). In this study, a few-shot learning technique occupies the model. The model is provided with a few examples from the target topic and integrated with the source topics in the gradual learning phase.

$$\mathcal{D}_{train} = (a \subset \mathcal{D}_s) \bigcup (b \subset \mathcal{D}_T)$$

Consequently, the model is anticipated to learn knowledge based on these examples, enhancing its capability to adapt to a new topic.

These components collectively engaged and provided a novel framework for gradual topic learning for Check-worthy Claim Detection, as illustrated in algorithm 1.

---

**Algorithm 1** Gradual Topic Learning (GTL)

---

**Require:** $\mathcal{D}_S \rightarrow$ Source domain,
$\quad\quad\quad\mathcal{D}_T \rightarrow$ Target domain,
$\quad\quad\quad\mathcal{M} \rightarrow$ Model,
$\quad\quad\quad\mathcal{T} \rightarrow$ number of Topics,
$\quad\quad\quad\mathcal{S} \rightarrow$ number of Stages

**Ensure:** Output includes the probability of check-worthiness for each $x_i^T \epsilon \mathcal{D}_T$
1: i=0, j=0
2: **for** for i in $\mathcal{T}$ **do**
3: $\quad$**for** for j in $\mathcal{S}$ **do**
4: $\quad\quad$**if** i < $\mathcal{S}$ **then**
5: $\quad\quad\quad Train\, \mathcal{M} \leftarrow (a \subset \mathcal{D}_s) \bigcup (b \subset \mathcal{D}_T)$
6: $\quad\quad$**else**
7: $\quad\quad\quad Train\, \mathcal{M} \leftarrow (b \subset \mathcal{D}_T)$
8: $\quad\quad$**end if**
9: $\quad$**end for**
10: **end for**
11: **Return** $MAP = \text{Test}(\mathcal{D}_T)$

---

## 4. Dataset

For this study, we experiment with tweets gathered from two datasets, CT20-AR and CT21-AR, which were released for shared tasks in CheckThat! 2020 and 2021 (Hasanain et al., 2020; Nakov et al., 2021b).

CT20-AR and CT21-AR are chosen as widely used claim detection datasets, where the collection of tweets provided are grouped into a variety of topics, all of them with tweets in the Arabic language. For instance, some of the topics in the dataset include social issues such as "Feminists", legal aspects like "The case of the Bidoon in Kuwait" political events such as "Boycotting countries and spreading rumours against Qatar", "Covid-19" as a global pandemic, religious views such as claims in the topic "Waseem Youssef", in addition to topics that are associated with different regions including MENA and USA.

The combined datasets comprise 8,100 tweets containing claims categorized into binary classes of checkworthiness, with (1) indicating *check-worthy* claims and (0) denoting *non-check-worthy* claims. The number of instances across the two classes in the dataset are imbalanced where there are 71.6% *non-check-worthy* claims against 28.4% *check-worthy* claims as shown in Figure 1.

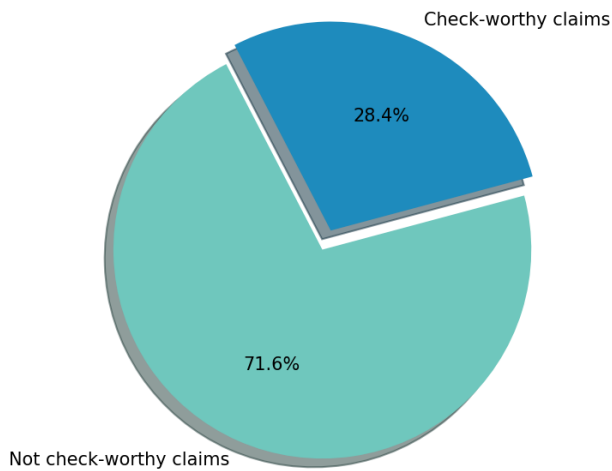

**Figure 1:** The overall proportion of worthy classes in CT20-AR and CT21-AR datasets.

***Topic distribution.*** The tweets cover a diverse range of topics. The CT20-AR dataset contains 15 topics with 500 tweets for each topic. The original CT21-AR dataset contains tweets that overlap with CT20-AR, and therefore this overlapping subset was removed, leaving the rest of the tweets. The remainder of the tweets include two new topics from CT21-AR, with 347 and 253 tweets. This originally led to 17 topics (15 from CT20-AR and 2 from CT21-AR); however, given that four of these topics were associated with COVID-19 and highly related to each other, we grouped these four topics into a single topic called COVID-19. This led to our final collection of 14 topics, which are listed in Table 1.

***Data preprocessing.*** To prepare the tweets for the experimental phase, we applied the Arabert's preprocessor function[1] to handle white spaces between words and digits and remove HTML markup, line breaks, extra spaces, unwanted characters, and emojis. Also, to include tokens as a replacement for some information such as [رابط] for a URL, [بريد] for an email address, and [مستخدم] for a username. Moreover, we customized and used another function to switch digits into the [رقم] token and to eliminate more unwanted symbols like punctuation and hash signs.

## 5. Methodology

In this section we describe the methodological framework employed to adapt a new topic for an Arabic check-worthy claim detection model. It includes an overview of the two proposed models: the GTL model, which mimics gradual learning, and the SGTL model utilizes a similarity-based strategy for sequencing the sentences of source and target domains in a rational order throughout the gradual learning stages. Next, we explain the evaluation metric and baseline we considered in our experiments.

---

[1]https://huggingface.co/aubmindlab/bert-base-arabert

| SN | TopicID | Title | Size |
|---|---|---|---|
| 1 | CT20-AR-01 | Deal of the century | 500 |
| 2 | CT20-AR-02 | Houthis in Yemen | 500 |
| 3 | CT20-AR-05 | Protests in Lebanon | 500 |
| 4 | CT20-AR-08 | Feminists | 500 |
| 5 | CT20-AR-10 | Waseem Youssef | 500 |
| 6 | CT20-AR-12 | Sudan and normalization | 500 |
| 7 | CT20-AR-14 | Events in Libya | 500 |
| 8 | CT20-AR-19 | Turkey's intervention in Syria | 500 |
| 9 | CT20-AR-23 | The case of the Bidoon in Kuwait | 500 |
| 10 | CT20-AR-27 | Algeria | 500 |
| 11 | CT20-AR-30 | Boycotting countries and spreading rumors against Qatar | 500 |
| 12 | COVID-19* | COVID-19 | 2k |
| 13 | CT21-AR-01 | Events in Gulf | 347 |
| 14 | CT21-AR-02 | Events in USA | 253 |

**Table 1**
Topics in CT20-AR and CT21-AR. COVID-19* encompasses four original topics in the same topic.

### 5.1. Gradual learning

Our proposed GTL and SGTL models are inspired by a combination of key insights from several studies. Building on the foundational concept in the literacy education, the gradual release of responsibility (GRR) model, one of the most often employed pedagogical frameworks, we have chosen (GRR) a concept that is contingent upon the principle of gradually releasing responsibility by the teacher to the learner, not by giving them total responsibility at once (Fisher and Frey, 2021). In addition, Bengio et al. (2009) introduced Gradual Fine-Tuning on NLP tasks, e.g. Dialogue State Tracking and Event Extraction. It relies on the idea of bootstrapping the model via gradually increasing the difficulty level of examples, which has been crucial in shaping our approach to topic adaptation. We hence design GTL and SGTL based on this intuition, enabling more effective adaptation of new topics.

### 5.2. GTL model

To implement the gradual fine-tuning method explained in the previous section 3, we follow a set of steps that we describe next.

***Train and test topic selection.*** 13 out of 14 topics in the dataset are grouped as source topics and one topic is deemed a target topic. Since the definition of gradual learning focuses on increasing the difficulty progressively, we assume the target topic is challenging for the model, which needs to be introduced gradually to the model during learning.

***Gradual sampling of target data.*** From the target topic, we use 200 samples as the ones that are gradually fed to the model for training, whereas the rest of the samples from the target topic are held out for testing. These 200 claims of the target topic are organized into smaller sets which are fed to the model for training in each of the stages; as we test our GTL model with different number of gradual training stages, the number of samples used in each stage is dependent on the number of stages. To determine how many samples to input in each stage, we use equation (1) to find a divisor first. Then, the number of samples to be allocated for each stage is calculated by applying equation (2); by multiplying the number of available claims by the number of the current stage and dividing it by the divisor. Recurrently, each step releases the remainder of the samples which are left for later stages, up until the last stage takes all of the remaining samples. Since the quantity of examples from the target topic used for learning is minimal, the final remaining calculation is added to the last subset to ensure the last stage has enough examples, considering the absence of source topics examples in the last stage. As a result, we created subsets that incrementally increased throughout the learning stages.

$$Y = \sum_{i=1}^{n} S_i + 2 \tag{1}$$

Where S is the number of subsets needed for the learning setting.

$$X = \frac{a \cdot i}{Y} \tag{2}$$

Where a holds a collection of claims, and i indicates the current stage.

***Provisioning the source domain data.*** For the source domain, ensuring the ideal distribution across different stages was uncertain to guarantee a coherent and progressive learning experience for the model. Thus, we planed to test two different apportionments for the source topics: (i) decrement gradient subsets and (ii) equivalent subsets, where the subsets spread through the stages except for the last stage, which is exclusive for learning the samples from the target topic.

***Provisioning the target domain data.*** The intention is to increase the portion of the target domain throughout learning gradually. Thus, we assume this requires decreasing the source domain. The performance for most stages was steady for this setting. This could be a potential reason for using an imbalanced dataset, where the check-worthy claims constantly decrease. Therefore, we have opted to test the maximum amount of the source domain as possible, which resulted in having equivalent subsets of the source domain.

In computing the decreasing subsets, we used the same process with equations (1) and (2), except the number of stages was reduced by one, as the last stage was eliminated for the source topics because its examples of source domains in the last stage should be equal to 0. Eventually, the outcomes were reversed to procure subsets of the source topics. Relating to determining the number of source topics examples required for the equivalent subsets, the total quantity of examples was divided by the number of stages-1. If any final remainder exists, it was added to the early stages.

***Merging source domain and target domain data for training.*** Finally, as illustrated in Figure 2, the training examples of the target topic are incorporated with the source topics in two different learning schemes throughout stages: A) decrements gradient subsets of the source domain with incremental subsets of the target domain; we refer to this scheme as **Dec-Inc**. B) equivalent subsets of the source domain with incremental subsets of the target domain; this scheme is noted as **Equ-Inc**.

***Classification stage.*** The input sentences are tokenized with Bert Fast before fine-tuning the AraBert model for the downstream classification task. To rank the check-worthy claims, we connect the last layer of the model to a softmax function in order to acquire the probability value for each prediction of check-worthy claims.

***Testing different numbers of stages.*** The aforementioned schemes were conducted in several groups of stages, including 2, 3, 6, 8, and 10 stages separately. Each group represents crucial learning steps for adapting the target domain, contributing to the comprehensive understanding of a new topic for the model, and boosting the ability to detect the check-worthy claim from the target domain in the testing phase.

## 5.3. SGTL model

The curriculum learning approach, which suggests arranging learning examples from simple to complicated, remains flexible as it can be developed using numerous criteria and distinct scheduling methodologies applicable at different levels, such as model, data, or task (Bengio et al., 2009; Soviany et al., 2022). Drawing upon this, in this paper, we propose a gradual learning composite with the concept of curriculum learning designed to expose the model to general topics and gradually shift the topics to become more similar to the target topic, gradually allowing it to emphasize the target domain examples.

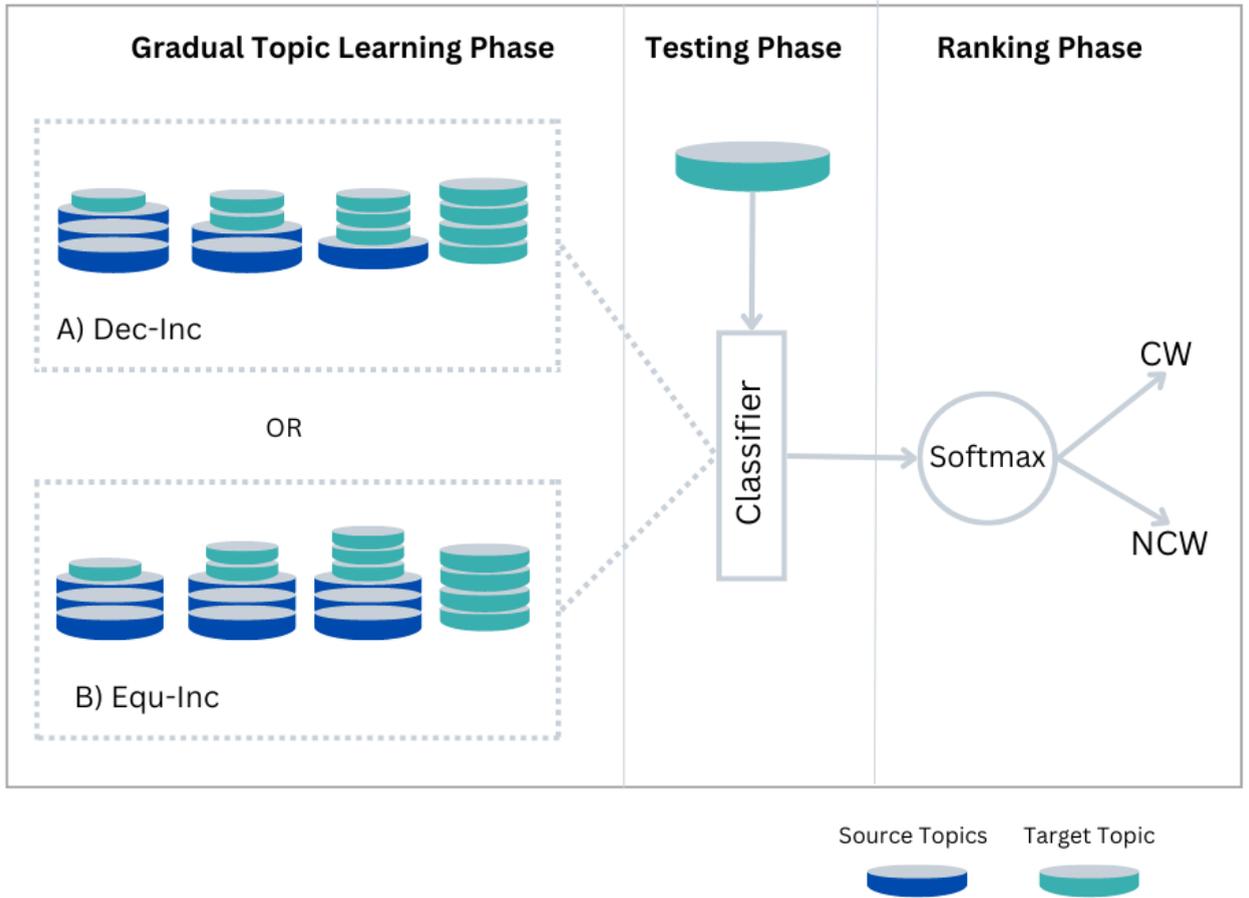

**Figure 2:** an in-depth overview of the GTL Model

In order to implement this strategy, it requires to estimate the similarity of claims in 13 topics of the source domain against the claims of the target domain. Hence, the appropriate metric according to this rule is the cosine similarity that computes the cosine of the angle between two vectors projected in a multi-dimensional space irrespective of the size of the topic, where each vector represents an array of word counts in a topic. The cosine similarity can be expressed as the following formula:

$$\cos(\theta) = \frac{\sum_{i=1}^{n} A_i B_i}{\sqrt{\sum_{i=1}^{n} A_i^2} \sqrt{\sum_{i=1}^{n} B_i^2}} \quad (3)$$

where $\mathbf{A} \cdot \mathbf{B}$ is the dot product of two vectors

Accordingly, the sequencing of the source topics in the SGTL model starts from dissimilar sentences to more homogeneous sentences to the target topics. Consequently, the ordered source domain via a similarity-based strategy is integrated with the target domain in two learning schemes: C) decrements gradient subsets of the source domain with incremental subsets of the target domain; this scheme is named **SBS_Dec-Inc**. D) equivalent subsets of the source domain with incremental subsets of the target domain; named **SBS_Equ-Inc**.

|                  | Baseline | s2     | s3     | s6     | s8     | s10    |
|------------------|----------|--------|--------|--------|--------|--------|
| **A_GTL_Dec-Inc**  | 0.6470   | 0.6932 | 0.6830 | 0.6787 | 0.6845 | 0.6623 |
| **B_GTL_Equ-Inc**  | 0.6470   | 0.6918 | 0.7140 | 0.7035 | 0.6591 | 0.6203 |
| **C_SGTL_Dec-Inc** | 0.6470   | 0.6668 | 0.6695 | 0.6926 | 0.6737 | 0.6080 |
| **D_SGTL_Equ-Inc** | 0.6470   | 0.6659 | 0.7046 | 0.7339 | 0.6627 | 0.6433 |

**Table 2**
Average performance of overall Results in GTL and SGTL models.

The overall logic, portioning and provisioning of source and target domain data with SGTL is similar to that of GTL, where the key difference is that with SGTL new data is provided by gradually increasing the degree of similarity and therefore by increasing the level of difficulty to enable the model to learn gradually more challenging topics towards the target topic.

### 5.4. Evaluation

We measure the performance of our models using the MAP (Mean Average Precision) metric to assess the model's capacity for identifying and ranking the most important check-worthy claims. The MAP scores are calculated from five separate runs for each scheme to get more steady and reliable performance estimations.

MAP is calculated as follows:

$$MAP = \frac{\sum_{c=1}^{C} \text{AveP}(c)}{C} \quad (4)$$

Where $C$ is the collection of claims in the test set, $c$ refers to each individual claim, and the equation for the Average Precision (AveP) is as follows:

$$AveP = \frac{\sum_{k=1}^{n} P(k) \ \text{rel}(k)}{|RC|} \quad (5)$$

Where $k$ refers to each position in the ranking of items produced, and $|RC|$ refers to the total number of claims.

The results are compared against the baseline (AraCWA) model without data augmentation, which is introduced as a state-of-the-art approach for detecting check-worthy claims across various topics. It leverages the few-shot learning technique by integrating 200 samples from the new topic alongside other topics during the fine-tuning of the AraBert model (Abumansour and Zubiaga, 2023).

## 6. Experimental Results

We next present the results of our experiments, beginning with the overall results of our two proposed models GTL and SGTL, followed by a detailed analysis of performance per topic and performing a comparison with the state-of-the-art baseline model AraCWA, and concluding by delving into the results with a qualitative analysis.

### 6.1. Overall results

Figure 3 and Table 2 present the average MAP scores for the four schemes, each conducted with different numbers of stages, i.e. 2, 3, 6, 8 and 10. In Figure 3, the two blocks on the left represent GTL, whereas the two blocks on the right represent SGTL.

Scanning the performance of the GTL model, which relies on gradual learning, specifically scheme (A) in Figure 3, which tests with (DEC-INC) data variation for domain and target topic, the results generally overcome the baseline, reaching the highest point when the model has two stages. In addition, it can be seen the steady pattern of results for stages 3, 6 and 8, with a marginal drop for ten stages close to the edge of the baseline.

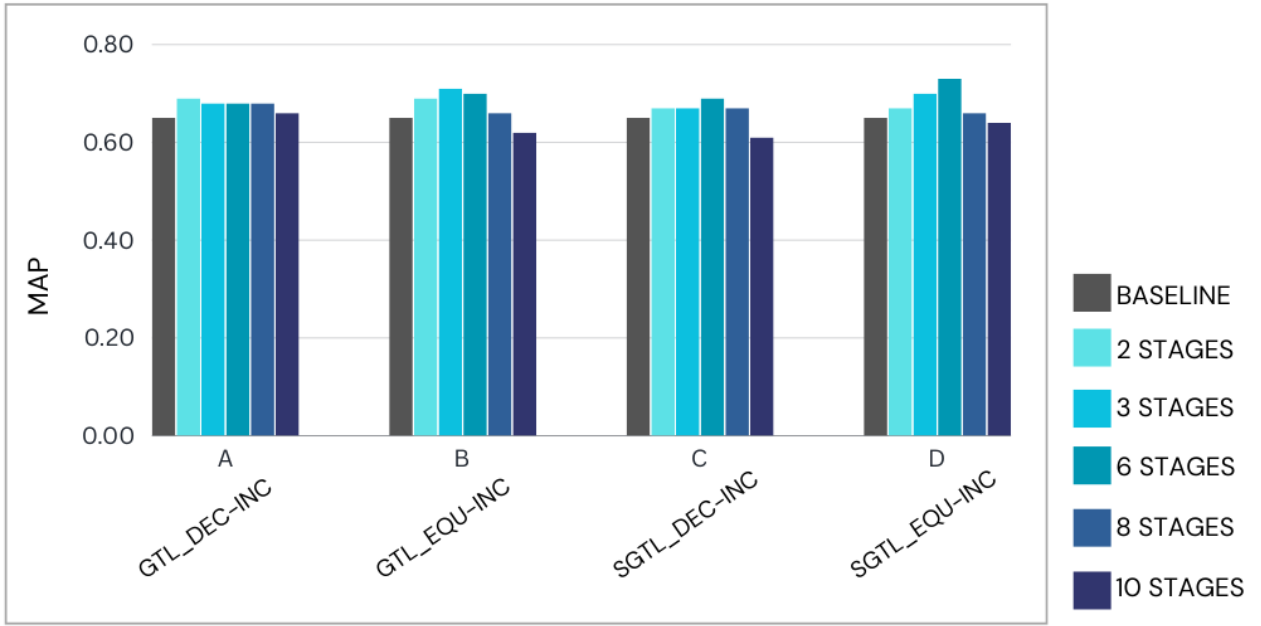

**Figure 3:** Average performance of overall results in GTL and SGTL models.

When the spreading of the source domain changed to be equal throughout the stages, giving a distribution of source and target topics as (EQU-INC) in the scheme (B), we observed growing performance compared to the baseline, which peaked in 3 stages and continuously declined when the learning length increased with the further stages. In the case of incorporating the similarity strategy with the "GTL model" to produce the enhanced "SGTL model," the source topics sequence through stages from different topics to similar topics to the target domain. What is interesting in comparison between the setting of both models with a similar distribution of source and target topics (DEC-INC), precisely scheme (A) for the GTL model and (C) for enhanced SGTL model, is the variability of results when the similarity strategy is activated, unlike the semi-firm outcomes in employ gradual learning solely. Therefore, this difference indicates that gradient learning for similar topics to the target topic has influenced the model to generalize the target topic.

Correspondingly, when contrasting the performance of engaging similarity strategy with gradual learning in the SGTL model against GTL model with gradual learning alone, specifically scheme (D) counter to scheme (B), where both hold the same source and target topics distribution as (EQU-INC), the outcome has denoted a clear improvement for scheme (D) as the results have continual growth and rose to a high point when the model learned for 6 stages within the same scheme (D) and among other schemes. Afterward, the results dramatically decreased at the following stages, e.g., 8, 10, respectively.

It is worth mentioning that when the model learns in two stages for both the GTL model and the SGTL model, the whole source domain is only seen in the first stage, along with samples from the target domain, while the second stage is solely limited to the target topic. Therefore, when the model has 2 learning stages, whether in the GTL model or the SGTL mode, schemes (A) and (B) are identical, and similarly, schemes (C) and (D). Nevertheless, their results closely mirror each other, asserting that gradual learning in multiple steps excels at adapting to new topics more skillfully than one or two learning steps.

These findings imply the importance of progressively employing topics similar to the target topic by utilizing a similar base strategy along with a gradual learning technique, as in the SGTL model. This leads to a close and in-depth understanding of check-worthy claims to be predicted from the new topic.

Moreover, we found the optimal number of learning stages in our study when the SGTL model has learning stages up to 6 since we noticed the performance constantly declined with a greater number of stages (i.e. 8 and 10 stages). This observation could indicate an association between data distribution and the count of learning stages, which warrants attention to avoid adverse results of the overlearning process.

| Topic ID    | AraCWA | SGTL-D6 | Imp. |
|-------------|--------|---------|------|
| CT20-AR-01  | 0.6883 | 0.7022  | 1%   |
| CT20-AR-02  | 0.6935 | 0.9231  | 23%  |
| CT20-AR-05  | 0.6002 | 0.9207  | **32%** |
| CT20-AR-08  | 0.3796 | 0.5439  | 16%  |
| CT20-AR-10  | 0.4660 | 0.6146  | 15%  |
| CT20-AR-12  | 0.8467 | 0.8778  | 3%   |
| CT20-AR-14  | 0.7354 | 0.7816  | 5%   |
| CT20-AR-19  | 0.8497 | 0.8945  | 4%   |
| CT20-AR-23  | 0.3723 | 0.3073  | -7%  |
| CT20-AR-27  | 0.6403 | 0.6392  | 0%   |
| CT20-AR-30  | 0.5730 | 0.7085  | 14%  |
| Covid-19    | 0.7101 | 0.7092  | 0%   |
| CT21-AR-01  | 0.6471 | 0.7717  | 12%  |
| CT21-AR-02  | 0.8554 | 0.88    | 2%   |
| Average     | 0.6470 | 0.7339  | 9%   |

**Table 3**
Results of SGTL Model in scheme D for 6 stages

### 6.2. Results per topic and baseline comparison

What is striking is the performance across topics for the most superior performance of models, the enhanced SGTL model that has 6 learning stages; it had effectively adapted the dominant of topics in our dataset when compared to the baseline of the AraCWA model as revealed in table 3, ranging positively between 1% and 32% except two topics spotted without achieving development which is "COVID-19" and CT20-AR-27 for "Algeria". In addition, the topic CT20-AR-23 for "The case of the Bidoon in Kuwait" has noticeably decreased by -7%. A probable reason for the case of "COVID-19" could be the rate of target samples to the amount of check-worthy claims to be predicted, e.g. (200:1800) where four original topics related to COVID-19 were combined, unlike the other topics that mostly have a ratio of (200:<300). Another possibility for the drop results is the difference between the target topic and the source topics to an extended level, such as the dialects and vocabularies the target topic holds.

### 6.3. Qualitative analysis

In addition to the quantitative results and analysis above, we also performed a qualitative analysis by looking at false negatives by the baseline model AraCWA which were correctly predicted by the SGTL-EQU-inc-s6 model. The purpose of this analysis is to ensure the effectiveness of the proposed SGTL-EQU-inc-s6 model in detecting check-worthy claims and assess the factors that could contribute to failure to perform with the problematic topics.

For implementing the qualitative analysis, we closely examined three scenarios for the topics performance result by SGTL-EQU-inc-s6, in particular, when the topic result significantly improved, the topic gnerlizability notable dropped, and when the topic revealed no enhancement result. We make the following key observations from this qualitative analysis:

- The first observation showed in best-performed topic CT20-AR-05 refers to topic "Protests in Lebanon" is the hashtag "Lebanon is rising up," or the word of "Lebanon" were present in most of false-negative of check-worthy claims e.g. 29 out of 31, this conveys that gradual learning helped the model to emphasis on the word "Lebanon" and the events related to the rising up in Lebanon specifically which lead to better generalizability and effectiveness of detection of check-worthy claims for the same topic.

- Second, while investigating on CT20-AR-23 for the topic "The case of the Bidoon in Kuwait" that has shown poor results with both AraCWA and SGTL-EQU-inc-s6 models, the false-negative of check-worthy claims revolve around the issue of the Societal group named "Bedoons" in particular. we have noticed there are some claims implicit with the writer's opinions which could casue to contextual misinterpretation for the claim by the model. Moreover, the word "Bidoon" was mentioned in all false-negative of check-worthy claims. In a word level, "Bidoon" or can be form as "Bedoon, or Bidun" is verbally equivalent to the word "without" which grammatically origins to "Doon" in Arabic "دُونَ" is a vogue adverb could be transferred to the noun or predicate form depending on the context (Alshamsan, 2016). In the context of the topic "The case of the Bidoon in Kuwait" the word "Bidoon" is an adverb used as a noun to name a social segment. Although the word "Bidoon" existed in all of false-negative check-worthy claim, the results of topic CT20-AR-23 has dramatic dropped by SGTL-EQU-inc-s6 model contrary to the impact of the word's occurrence for the word "Lebanon" in topic CT20-AR-05 which is an explicit noun used devoid of semantic change, thus can be explained by a reason of semantic inflection of the word which could change the meaning of the context. Thus, the first finding of having high occurrence of a specific word in a topic helps to generalize this topic by gradual learning would have been more valid if a wider range of contextual and semantic perspectives considered.

- Lastly, CT20-AR-27 for the topic of "Algeria" did not improve by the SGTL-EQU-inc-s6 model and yielded the same result as with the baseline model AraCWA. We found the false-negative check-worthy claims for CT20-AR-27 consist of Multi-Subtopics such as political, sports, economic, social, and legal issues related to homosexuals and COVID-19 cases in Algeria, unlike the best-enhanced topic about the "protests in Lebanon" which focused on a specific event in Lebanon. Therefore, the diversity of topics could confuse the model and mislead the classification of many claims, as previously proved in the study of the detection of check-worthy claims across topics (Abumansour and Zubiaga, 2023).

## 7. Conclusions

In this study, we investigated the impact of gradual learning on adaptation to a new topic for a check-worthy claim detection model for the Arabic language. Our study indicates that the improved SGTL model with a well-implemented similarity-based strategy significantly outperforms in generalizability across topics. Specifically, we observed a notable enhancement when the SGTL model learned about the new topic in six stages while engaging and retaining key concepts of checky-worthy claims. These results suggest that Similarity-driven Gradual Topic Learning can be a crucial strategy in designing a Check-worthy Claim Detection, providing a consistent learning experience throughout stages that support long-term topic adaptation achievement. This has outstanding implications for Automated Fact-checking systems.

Our experiment identifies 6 stages as the optimal learning stages in the claim detection task with the best model performance SGTL-EQU-INC, whereas (Bengio et al., 2009) their task indicates 4 as ideal stages. Therefore, the optimal number of stages expended in future experiments could vary depending on several factors, including the size and distribution of the data. Additionally, adaptation challenges with some topics could be referred to other factors, such as semantics, linguistics, or how branched the topic is, which call for further exploration in the future. Nevertheless, gradual learning significantly outperforms single learning. Furthermore, the SGTL-EQU-INC model demonstrates the capacity to employ a similarity-driven strategy to adapt to new topics efficiently. To the best of our knowledge, this is the most recent technique developed for detecting check-worthy claims for AFC systems.

While our work advances research in the ability to improve claim detection across new, unseen topics, it is not without limitations. The first main drawback of our study is the relatively small sample size of the target topic unified with the source domain for gradual learning, which could restrict the generalization of the new topic. Second, rigorous data preparation requires concentration and ample time to identify the target topic and prepare the sets of the source domain, target domain, and samples of the target domain individually for each new topic. Lastly, this study handled Arabic check-worthy tweets exclusively. The outcome may not be standardized to other sizes, genera, or the language of data, thus demanding future research extension.

## CRediT authorship contribution statement

**Amani S. Abumansour:** . **Arkaitz Zubiaga:** .

# References

Aarnes, P. R., Setty, V., and Galuščáková, P. (2024). Iai group at checkthat! 2024: Transformer models and data augmentation for checkworthy claim detection. *arXiv preprint arXiv:2408.01118*.

Abumansour, A. S. and Zubiaga, A. (2023). Check-worthy claim detection across topics for automated fact-checking. *PeerJ Computer Science*, 9:e1365.

Abumansour, A. S., Zubiaga, A., et al. (2021). Qmul-sds at checkthat! 2021: enriching pre-trained language models for the estimation of check-worthiness of arabic tweets. *CLEF*.

Alam, F., Barrón-Cedeño, A., Cheema, G. S., Shahi, G. K., Hakimov, S., Hasanain, M., Li, C., Míguez, R., Mubarak, H., Zaghouani, W., et al. (2023). Overview of the clef-2023 checkthat! lab task 1 on check-worthiness of multimodal and multigenre content. *Working Notes of CLEF 2023—Conference and Labs of the Evaluation Forum*.

Alshamsan, A. I. (2016). أحكام (دون) ودلالاتها السياقية. *El-Harakah*, 18(2):223–242.

Antoun, W., Baly, F., and Hajj, H. (2020). Arabert: Transformer-based model for arabic language understanding. In *Proceedings of the 4th Workshop on Open-Source Arabic Corpora and Processing Tools, with a Shared Task on Offensive Language Detection*, pages 9–15.

Atanasova, P., Barron-Cedeno, A., Elsayed, T., Suwaileh, R., Zaghouani, W., Kyuchukov, S., Martino, G. D. S., and Nakov, P. (2018). Overview of the clef-2018 checkthat! lab on automatic identification and verification of political claims. task 1: Check-worthiness. *arXiv preprint arXiv:1808.05542*.

Atanasova, P., Nakov, P., Màrquez, L., Barrón-Cedeño, A., Karadzhov, G., Mihaylova, T., Mohtarami, M., and Glass, J. (2019). Automatic fact-checking using context and discourse information. *Journal of Data and Information Quality (JDIQ)*, 11(3):1–27.

Aziz, A., Hossain, M. A., and Chy, A. N. (2023). Csecu-dsg at checkthat!-2023: Transformer-based fusion approach for multimodal and multigenre check-worthiness. In *CLEF (Working Notes)*, pages 279–288.

Babakar, M. and Moy, W. (2016). The State of Automated Factchecking. Technical report, Full Fact, London, UK.

Bengio, Y., Louradour, J., Collobert, R., and Weston, J. (2009). Curriculum learning. In *Proceedings of the 26th annual international conference on machine learning*, pages 41–48.

Brown, T. B. (2020). Language models are few-shot learners. *arXiv preprint arXiv:2005.14165*.

Du, M., Gollapalli, S. D., and Ng, S.-K. (2022). Nus-ids at checkthat! 2022: Identifying check-worthiness of tweets using checkthat5. In *CLEF (Working Notes)*, pages 468–477.

Elsayed, T., Nakov, P., Barrón-Cedeno, A., Hasanain, M., Suwaileh, R., Da San Martino, G., and Atanasova, P. (2019). Overview of the clef-2019 checkthat! lab: automatic identification and verification of claims. In *Experimental IR Meets Multilinguality, Multimodality, and Interaction: 10th International Conference of the CLEF Association, CLEF 2019, Lugano, Switzerland, September 9–12, 2019, Proceedings 10*, pages 301–321. Springer.

Fernandez, M. and Alani, H. (2018). Online misinformation: Challenges and future directions. In *Companion proceedings of the the web conference 2018*, pages 595–602.

Fisher, D. and Frey, N. (2021). *Better learning through structured teaching: A framework for the gradual release of responsibility.* ASCD.

Hasanain, M., Haouari, F., Suwaileh, R., Ali, Z. S., Hamdan, B., Elsayed, T., Barrón-Cedeño, A., Martino, G. D. S., and Nakov, P. (2020). Overview of checkthat! 2020i arabic: Automatic identification and verification of claims in social media. In *Proceedings of CLEF*.

Hasanain, M., Suwaileh, R., Weering, S., Li, C., Caselli, T., Zaghouani, W., Barrón-Cedeño, A., Nakov, P., and Alam, F. (2024). Overview of the clef-2024 checkthat! lab task 1 on check-worthiness estimation of multigenre content. *Working Notes of CLEF*.

Hassan, N., Arslan, F., Li, C., and Tremayne, M. (2017a). Toward automated fact-checking: Detecting check-worthy factual claims by claimbuster. In *Proceedings of the 23rd ACM SIGKDD international conference on knowledge discovery and data mining*, pages 1803–1812.

Hassan, N., Zhang, G., Arslan, F., Caraballo, J., Jimenez, D., Gawsane, S., Hasan, S., Joseph, M., Kulkarni, A., Nayak, A. K., et al. (2017b). Claimbuster: The first-ever end-to-end fact-checking system. *Proceedings of the VLDB Endowment*, 10(12):1945–1948.

Hou, B., Chen, Q., Shen, J., Liu, X., Zhong, P., Wang, Y., Chen, Z., and Li, Z. (2019). Gradual machine learning for entity resolution. In *The World Wide Web Conference*, pages 3526–3530.

Jaradat, I., Gencheva, P., Barrón-Cedeño, A., Màrquez, L., and Nakov, P. (2018). ClaimRank: Detecting check-worthy claims in Arabic and English. In *Proceedings of the 2018 Conference of the North American Chapter of the Association for Computational Linguistics: Demonstrations*, pages 26–30, New Orleans, Louisiana. Association for Computational Linguistics.

Konstantinovskiy, L., Price, O., Babakar, M., and Zubiaga, A. (2021). Toward automated factchecking: Developing an annotation schema and benchmark for consistent automated claim detection. *Digital threats: research and practice*, 2(2):1–16.

Nakov, P., Barrón-Cedeño, A., Da San Martino, G., Alam, F., Struß, J. M., Mandl, T., Míguez, R., Caselli, T., Kutlu, M., Zaghouani, W., et al. (2022). The clef-2022 checkthat! lab on fighting the covid-19 infodemic and fake news detection. In *European Conference on Information Retrieval*, pages 416–428. Springer.

Nakov, P., Corney, D., Hasanain, M., Alam, F., Elsayed, T., Barrón-Cedeño, A., Papotti, P., Shaar, S., Da San Martino, G., et al. (2021a). Automated fact-checking for assisting human fact-checkers. In *Proceedings of the Thirtieth International Joint Conference onArtificial Intelligence,˜IJCAI-21`*, pages 4551–4558. International Joint Conferences on Artificial Intelligence

Organization.

Nakov, P., Da San Martino, G., Elsayed, T., Barrón-Cedeno, A., Míguez, R., Shaar, S., Alam, F., Haouari, F., Hasanain, M., Babulkov, N., Nikolov, A., Shahi, G. K., Struß, J. M., and Mandl, T. (2021b). The clef-2021 checkthat! lab on detecting check-worthy claims, previously fact-checked claims, and fake news. In *European Conference on Information Retrieval*, pages 639–649. Springer.

Panchendrarajan, R. and Zubiaga, A. (2024). Claim detection for automated fact-checking: A survey on monolingual, multilingual and cross-lingual research. *Natural Language Processing Journal*, 7:100066.

Sadouk, H. T., Sebbak, F., and Zekiri, H. E. (2023). Es-vrai at checkthat! 2023: Analyzing checkworthiness in multimodal and multigenre. *CLEF*.

Shaar, S., Hasanain, M., Hamdan, B., Ali, Z. S., Haouari, F., Alex Nikolov, M. K., Yavuz Selim Kartal, F. A., Da San Martino, G., Barrón-Cedeño, A., Míguez, R., Elsayed, T., and Nakov, P. (2021). Overview of the CLEF-2021 CheckThat! lab task 1 on check-worthiness estimation in tweets and political debates. In *Working Notes of CLEF 2021—Conference and Labs of the Evaluation Forum*, CLEF '2021, Bucharest, Romania (online).

Singhal, P., Walambe, R., Ramanna, S., and Kotecha, K. (2023). Domain adaptation: challenges, methods, datasets, and applications. *IEEE access*, 11:6973–7020.

Soviany, P., Ionescu, R. T., Rota, P., and Sebe, N. (2022). Curriculum learning: A survey. *International Journal of Computer Vision*, 130(6):1526–1565.

Williams, E., Rodrigues, P., and Novak, V. (2020). Accenture at checkthat! 2020: If you say so: Post-hoc fact-checking of claims using transformer-based models. *arXiv preprint arXiv:2009.02431*.

Xu, H., Ebner, S., Yarmohammadi, M., White, A. S., Van Durme, B., and Murray, K. (2021). Gradual fine-tuning for low-resource domain adaptation. In Ben-David, E., Cohen, S., McDonald, R., Plank, B., Reichart, R., Rotman, G., and Ziser, Y., editors, *Proceedings of the Second Workshop on Domain Adaptation for NLP*, pages 214–221, Kyiv, Ukraine. Association for Computational Linguistics.

Zeng, X., Abumansour, A. S., and Zubiaga, A. (2021). Automated fact-checking: A survey. *Language and Linguistics Compass*, 15(10):e12438.

Zhang, X., Shapiro, P., Kumar, G., McNamee, P., Carpuat, M., and Duh, K. (2019). Curriculum learning for domain adaptation in neural machine translation. *arXiv preprint arXiv:1905.05816*.

Zubiaga, A. (2019). Mining social media for newsgathering: A review. *Online Social Networks and Media*, 13:100049.